# A HUMAN-CENTRIC APPROACH TO GROUP-BASED CONTEXT-AWARENESS

Nasser Ghadiri[1] , Ahmad Baraani-Dastjerdi[2,*], Nasser Ghasem-Aghaee[3], Mohammad A. Nematbakhsh[4]

[1]Department of Computer Engineering, University of Isfahan, Isfahan, Iran
ghadiri@eng.ui.ac.ir
[2]Department of Computer Engineering, University of Isfahan, Isfahan, Iran
ahmadb@eng.ui.ac.ir
[3]Department of Computer Engineering, University of Isfahan, Isfahan, Iran
aghaee@eng.ui.ac.ir
[4]Department of Computer Engineering, University of Isfahan, Isfahan, Iran
nematbakhsh@eng.ui.ac.ir

## ABSTRACT

*The emerging need for qualitative approaches in context-aware information processing calls for proper modelling of context information and efficient handling of its inherent uncertainty resulted from human interpretation and usage. Many of the current approaches to context-awareness either lack a solid theoretical basis for modelling or ignore important requirements such as modularity, high-order uncertainty management and group-based context-awareness. Therefore, their real-world application and extendibility remains limited. In this paper, we present f-Context as a service-based context-awareness framework, based on language-action perspective (LAP) theory for modelling. Then we identify some of the complex, informational parts of context which contain high-order uncertainties due to differences between members of the group in defining them. An agent-based perceptual computer architecture is proposed for implementing f-Context that uses computing with words (CWW) for handling uncertainty. The feasibility of f-Context is analyzed using a realistic scenario involving a group of mobile users. We believe that the proposed approach can open the door to future research on context-awareness by offering a theoretical foundation based on human communication, and a service-based layered architecture which exploits CWW for context-aware, group-based and platform-independent access to information systems.*

## 1. INTRODUCTION

As envisioned by Mark Weiser in [1], ubiquitous information access for people requires the computing technology to become as simple as older technologies like printing. After almost two decades of research and tremendous amount of work on technology enhancements in mobile computing, still a large gap exists between today's computing technologies and the user's environment, requirements and expectations in utilizing such technologies. The number of available mobile devices was almost equal to half of the world population by the end of 2008. Such rapid prevalence of mobile devices is shaping our new lifestyles, and more human-centric approaches will be required to support our daily activities and future mobile working environments [2]. Context-awareness plays a key role in adapting the information systems to real needs of the user. The idea has a rather long history. The artificial intelligence society was

---

* Corresponding author

                                                      47

criticized in 1980's for ignoring the differences in situations where software was used [3]. In the last decade, a large body of research on context-awareness has been performed. Two recent literature reviews and classifications are done by Hong, et al in [4] and by Balduf, et al in [5]. To the best knowledge of us, most approaches have one of the following shortcomings. First, current approaches lack a sound theoretical and architectural basis to fulfil the need for establishing an extensible and platform-independent framework. Second, most approaches are weak in terms of accommodating human-centric, uncertainty-aware design requirements for adapting to real-world applications. Even ontological approaches which are suggested for context-awareness as well as other application areas have a major problem namely symbol grounding (ref to Gardenfors). More details about some important challenges of context-awareness is given in Section 0.

To solve the first problem, we suggest using the theory of Language Action Perspective (LAP) to define the position of context, and a layered, service-based architecture for extensibility and platform-independence. As Winograd stated in [6], service-based architectures of context provide this key feature. Based on the LAP theory, we can obtain a set of design methods to model programs that are closer to what people need [7]. The benefits of this theory-based approach in service-oriented computing, and the weakness of existing research in exploiting such theories is discussed in detail by [8]. A LAP-inspired framework for web services is proposed by [9] which we will use to establish position of context in different layers. LAP has its roots in speech act theory and defines a perspective in which people use language as a communication method to perform actions. The key difference of such theories with traditional information-process thinking and modelling is the order in which *communication* and *information processing* happens [10]. However, analyzing communication is not as straightforward as information and is highly context-dependent [3]. An important aspect of this complexity is due to the fact that communication is primarily done by humans, which contains inherent *uncertainties* that should be dealt with. As mentioned above, this is the second problem of current context-awareness.

Therefore, the second problem, the problem of uncertainty and human-centric design, is not directly addressed by the LAP theory. Based on LAP, we can only define *what* requirements should be considered for a viable context-aware design. The next key question here is *how* to exploit the benefits of the proposed LAP-based framework. We need to model the context, with its inherent complex uncertainty, in a way that it becomes understandable and manageable by such a high level theoretical approach. We will show that a good solution to this second problem is moving from traditional methods of handling *data*, toward handling *perceptions*, for which the name "computing with words" (CWW) was coined by Zadeh in [11]. CWW makes it possible to perform qualitative reasoning tasks using advanced fuzzy logic theories.   In this paper, we will focus on spatial, temporal and human dimensions of context, which contain more complex types of uncertainty, and have attracted less research interest before. We will show that a major source of uncertainty is the different interpretations of a word between members of a *group* or a community of users. Therefore, a mechanism for involving the group members in defining and redefining these context-related words which are used later in group decision making rules will be introduced.

The main contributions of this paper are:

1. Defining the position of context in a larger theory of LAP for providing answers to some crucial questions of *what* type about context-awareness

2. Offering a novel conceptual view of context which separates the simple, raw context parts from rich, informational parts and delves deeper into three of such parts namely spatial, temporal and psychological domains

3. Designing a layered architecture for providing platform-independent, autonomic context services and verifying its feasibility using a sample scenario

The rest of the paper is organized as follows. In Section 2 we give some motivation for the cogency of handling uncertainty in context-aware systems. Location-awareness, temporal-awareness and psychological need-awareness are special cases of context-awareness. Hence, our approach to uncertainty in context in subsequent sections will be multidisciplinary. Section 3 presents the main theoretical foundations which support our design including the LAP theory, supported by CWW, uncertain topological relationships in the spatial part of context and the Existence, Relatedness, Growth (ERG) theory of human needs. Then we present our conceptual view of context, discriminating between the raw data parts, and more complex information parts which require high-order uncertainty management. In Section 4 we present our architectural design, f-Context, as an agent-based perceptual computer and provides various services which can be used by context-aware applications. Verification of our design using a real world scenario is presented in Section 5, followed by related works in Section 6. Section 7 concludes the paper and points to some possible future research.

## 2. MOTIVATIONS AND RESEARCH CHALLENGES

In this section we present some important challenges of search and access to information with different perspectives to emphasize both context-awareness and finding methods for handling its uncertainty.

*The problem of current context-oblivious search:* Although today's search engines are shown to be powerful in finding information, they are still far from ease-of-use in a qualitative and question-answering style, usable by non-expert people [12]. More human-centric approaches are required, to proactively percept user's needs, desires and context in general, for overcoming current problems of less-successful searches which return overabundances of irrelevant information in some cases. Inspecting some real world search data from two aspects of spatial [13] and human needs [14] show that a very small portion (less than %3) of queries do explicitly contain what a user is looking for. Moreover, almost half of the results were not clicked by the user at all, i.e. no interesting information was returned. This is partially due to the lack of user's context information when searching. Even a small enhancement in context-awareness, will make the search more efficient. Context-awareness can also improve searching for information services. In [15] three parts of context information, i.e. temporal, spatial and user activity parts are considered to be a time-saving facility for service discovery. However, they believe that much research is still needed for utilizing such imprecise context information.

*Mobility, heterogeneity, and dynamic information:* In addition to the Internet-based and non-mobile applications, the last-mentioned problem of search and access information exists and is even more challenging in ubiquitous information access environments. It can be considered as a generalization of the discussed problem, which also adds the complexity of heterogeneity and dynamic nature of information sources [16]. Such environments are evolving the context from a pure informational form toward interactional forms, in which context is arising from activities, communication and interaction between users [17]. This new perspective of context-awareness will require more rich modelling and more efficient reasoning in such uncertain and changing environments.

*The role of group and high-order uncertainty:* The increased role of groups can be seen in today's community-based activities like search and access to information especially for mobile users. As shown by [18], extending the context from individual context to group context allows us to build a context model that is more human-centric, and it "…resembles the way people adapt their behaviour in groups. A collaboratively agreed context may be more reliable than that of a single device.". Several real-world application domains can utilize the benefits of a sound group-based context architecture. A promising field is group-based services in mobile commerce. There are real problems in conducting e-commerce transactions between mobile users like m-auctions. Context-awareness can augment the platforms which are designed for supporting such services [19]. More specifically, the *spatial* part of context can help to overcome some mobility limitations and problems. For instance, learning the persisting information about the mobility patterns of a user in long-term context knowledgebase will help the system to decide about short periods of disconnection. However, the spatial context contains high-order uncertainty [20], and new methods are necessary to manage it and to make qualitative reasoning tasks more efficient, especially in group mode. Another application domain is Group Information Management (GIM), an extension of Personal Information Management (PIM) systems. These disciplines aim at systematic orderliness for overcoming the problem of individual and group information overloads [21]. We believe that the long-term personal and group contexts which contain our historical context information, criteria and activities based on our usage of services, can be considered as personal and group knowledge repositories respectively. It should be emphasized that the group itself is a source of high-order uncertainties due to the fact that "words mean different thing to different people", as noted by Mendel [22].

*Security, privacy and trust:* The trade-off between trust and information sharing and exposure which is usually required in service-oriented computing, can be enhanced by contextualization. For instance, we can select to reveal the requested parts of personal context, only to those mobile users and/or service providers who are *spatially* or (in a more generalized form) are *contextually* near to us, thus preventing unnecessary revelation of our information and tasks.

*The trend toward more human-centric, qualitative reasoning:* In artificial intelligence and modelling human's way of reasoning, qualitative information is preferred as it provides more meaning than quantitative data [23]. For example, in spatial reasoning, knowing about the relative position of a moving mobile user to nearby friends can be more useful than knowing the exact numeric coordinates of a large set of trajectory points. People also exchange such qualitative information when communicating about a fact. Pragmatic theories of language use, e.g. speech act, focus on user communication rather than pure information, and complex human communication highly depends on context [3]. Therefore, we also need powerful methods to handle the qualitative parts of context such as spatial and temporal parts. It is shown that context –awareness requires a deeper insight into human needs [24]. In this paper, we will also try to handle the contextual human needs and motivations, which exist behind our communicative acts, by using a psychological theory of human needs.

*Long-term vs. transient short-term context:* While many of current approaches to context-awareness focus on last-minute information about user's context, some applications require long-term context for high-quality decisions. For example, an m-commerce application may use long-term information about user's location to estimate the reliability of transactions performed by this user. Therefore, both long- and short-term aspects of context should be considered in future designs to fulfil such requirements.

Our proposed f-Context framework is designed as a starting point for overcoming such problems which have not been addressed by current context-aware approaches. Our approach will be based on several theories including the LAP theory, which helps to model the context by focusing on human communication, and the theory of CWW which is based on type-2 fuzzy logic and helps in handling high-order uncertainty which exists in that communication between group members, and some domain-related theories for spatial, temporal and other parts of context.

# 3. THE PROPOSED F-CONTEXT FRAMEWORK

In this section we introduce the LAP-inspired framework which is used to position our proposed f-Context in different layers of a service-based architecture. Then our conceptual view of context is presented, which separates simple, raw quantitative parts of context from more rich, qualitative sources of context information. We also give a brief introduction to spatial, temporal and psychological human need domains which are considered as three important roots of high-order uncertainty.

## 3.1. The LAP-inspired framework and the position of f-Context

As discussed in Section 2, there is a trend from static, informational view of context toward a more dynamic, interactional view. The LAP theory introduced in [7] is based on speech act theory, and can help to provide such a view. LAP is used by [9] as the basis for a novel layered framework to organize services-based access to information. The benefits of using this approach is discussed in [8]. We will use this LAP-inspired framework for our work.

The LAP-inspired framework of [9] consists of 11 *layers* which are grouped into 3 main *levels*. The first lower *communication platform* level is about '*how*' communicative acts are performed, and consists of 3 layers, mainly concerned with basic channel operations, messaging and so on. These layers do not have much dependency on context. Most of the existing standards for service-based access to information fully support this layer. The second level is *communicative act*, which concerns '*what*' aspects of communication, and contains 4 layers for capability exposure and search, proposal and contracting. All service search and discovery approaches fit into this level. We believe that dependency on context starts at this level for short-term, session-specific operations like discovering information services. The third level, *rational discourse*, concerns the '*why*' aspects of communication. This level has a higher context-dependency for long-term processes such as relationship management between an information system and its users. The degree of context-dependency increases in upper layers, due to the essence of each layer in answering the above questions and with increasing user involvement compared to lower layers.

Where does f-Context or the context in general, fit into this LAP-inspired model? The answer is somewhat uncertain itself. As described above, several layers can utilize the context-awareness provided by f-Context. So it is would be a good idea to consider the context as an integrated inter-layer service, as depicted in Figure 1. The *relationship management* and *capability search* layers have pivotal role in our design for both managing and using long-term and short-term context, respectively. We describe these two layers briefly.

The *Relationship management* layer is responsible for managing long-term relations between, say, a company and its customers in a business perspective [25]. The survey by [9] shows that none of the aforementioned standards of service-orientation offer relationship management facilities. The presence of a persistent long-term context in f-Context will help such operations to be handled properly, as will be discussed in our design and the sample scenario.

The *capability search* layer, which has attracted a large body of research in syntactic and semantic service discovery, will utilize the short-term and session-specific capabilities of f-Context to provide improved search and discovery facilities, as close to personal and/or group context as possible.

This general approach can be applied to other types of services as well, including mobile services in which the service provider and/or service consumer are moving [26]. Due to dynamically changing location of users and services, mobile services require more *spatial* context-awareness, which surrounds the mobile user and mobile service provider, and will be discussed in more detail later. In addition, *temporal* context-awareness will help a group to make efficient decisions about surrounding events and services.

## 3.2. Realizing the LAP using CWW based on type-2 fuzzy logic

A major barrier for using the LAP theory is lack of an implementation method. In LAP, the communication between people is done using words from a context-dependent vocabulary. Because human is engaged in this communication, there are high-order uncertainties associated with words. Therefore, the selected method should be mathematically powerful to cope with high-order uncertainty of group communication. On the other hand, it should be simple enough to be used by group members who communicate using natural language. In other words, we have two different worlds with a large gap between them. Fuzzy sets have a long history of solving this problem [27, 28] by playing the role of a bridge between the two worlds using the concept of linguistic variables. For each linguistic variable, like *near* and *far* in spatial relations, a membership function can be defined and used when reasoning about distance.

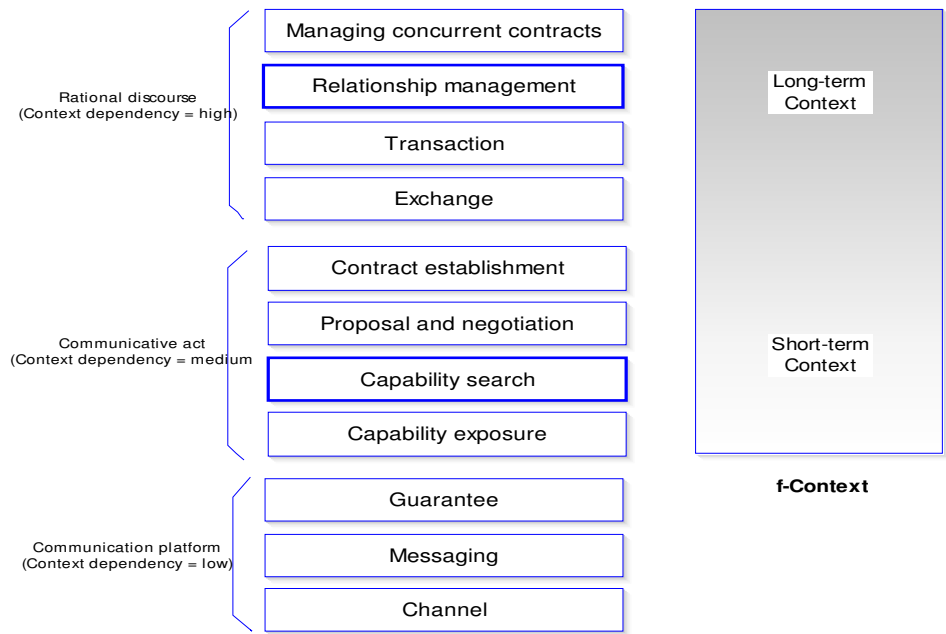

Figure 1: Position of our f-Context in a LAP-inspired framework.

However, Mendel in [29] uses Karl Popper's falsification theory to prove that type-2 fuzzy sets are better than classical (type-1) fuzzy sets for modelling words. He argues that type-1 fuzzy sets are based on membership functions that are crisp themselves, while people are uncertain about the exact form of the membership functions. CWW based on type-2 fuzzy sets allows this uncertainty to be managed, by allowing the membership functions to have an extra degree of freedom for handling the differences between people in defining the meaning of each word.

Therefore we propose to implement the LAP-based framework, by using CWW based on type-2 fuzzy logic. By using this method we can handle perceptions (words), process them, and return the result using words. To establish this CWW-based approach, first we need to present our model of context, its important information domains, and the inherent uncertainties associated with the domains. Then the details of our proposed f-Context architecture which exploits CWW to for context-awareness will be presented in Section 4.

## 3.3. A conceptual view of context

Our conceptual view of context is shown in Figure 2. It consists of simple, raw, sensor-supplied numeric data parts shown in the left side, and more complex information parts shown on the right side. The simple numeric data parts of context may represent well-defined categorical data, such as vehicle type and device type. There is no sensible uncertainty associated with this type of data. They can also represent *quantitative* data, which are continuous variables like network bandwidth, level of light, temperature and so on. They can contain simple type of uncertainty, due to randomness, measurement errors and so on. However, handling this type of uncertainty is rather straightforward by using fuzzy logic and probabilistic methods and is already discussed in many sources, including but not limited to [20, 30-33].

In this section we focus on complex parts shown in the right side of Figure 2. For a human-centric model of context, we are facing with modelling and reasoning methods of *qualitative* data, i.e. contextual information, which can be resulted from generalizing original quantitative data and exist in complex parts of context, e.g. the spatial context. They are also categorical in nature, i.e. their values are selected from a limited number of values, but their boundaries cannot be exactly defined. In the following, we will introduce three important types of such contextual information with more detail.

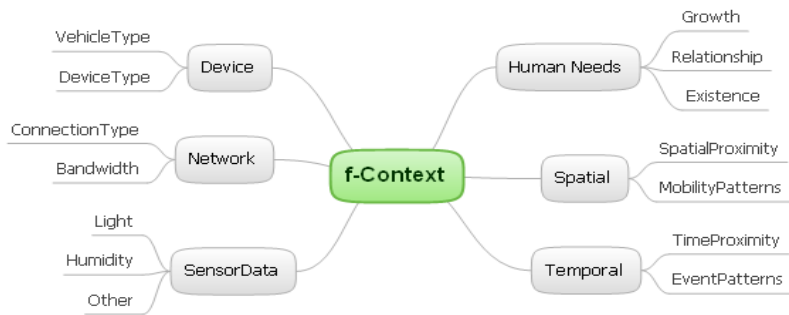

Figure 2. Our conceptual view of context elements (partial). Left part: Simple, raw context elements. Right part: Ripe context elements with high-order uncertainties.

### 3.3.1. The spatial part of context

The spatial part of context reflects information about the location of user or service. This part of context is treated in literature with various approaches ranging from quantitative Global Positioning System (GPS)-based approaches to more complex spatial and spatio-temporal proximity relationships. The quantitative approaches are computationally intensive and require multiple spatial joins even for the simplest uncertain queries. What we need is a qualitative approach for uncertain perceptions by human and agents. Perception of spatial configurations includes identification of *topological relations* such as overlapping and touching between regions [34]. To give a better understanding of spatial uncertainty for qualitative reasoning tasks, we provide an example, which will be also used in our scenario in Section 0. As shown in Figure 3, suppose that we need to decide whether a mobile user or service provider who is located near the 'Eiffel tower' is *next to* the river or not.

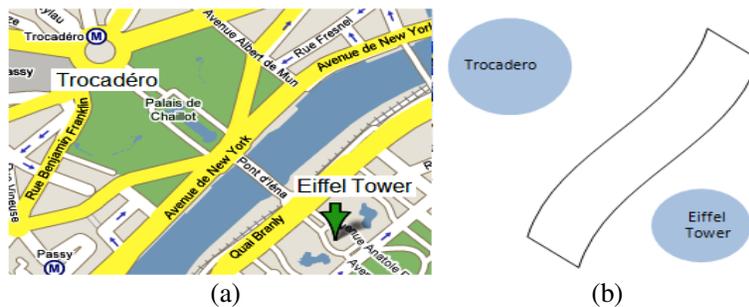

Figure 3. Spatial uncertainty: (a) Google maps image, (b) topological representation of the image

This is a qualitative uncertain decision and is difficult to handle with normal reasoning techniques. In terms of classic topology, the tower and the river do not precisely overlap, but we intuitively can see from the figure that they are *near* to each other. Based on this map, the distance between the Eiffel tower and the Saint river is about 200 meters, while the distance from Trocadero to the river is about 500 meters. If we ask several people "Is the Eiffel tower is located *next to* the river" and "What about Trocadero?", we may get different responses from different people (inter-user spatial uncertainty). This uncertainty, if not properly handled, may lead to conflict in group and blocking of group activities. So we must find a group-acceptable method to manage it based on group consensus, i.e. using type-2 fuzzy sets as will be proposed in the f-Context architecture in Section 4.

### 3.3.2. The temporal part of context

Everything that is related to past, present and future situations of a user or service, can be modelled by temporal part of context. When combined with spatial context, the resulting spatio-temporal context is useful in many reasoning algorithms about context. Context reasoning services of this type are also important for the relationship management layer in LAP, which deal with the long-term past and future of context. It should be noted that qualitative reasoning about time is also uncertain and complex in some cases. The results of temporal reasoning by human are not always formally verifiable. For example, if you ask a taxi driver to take you to a specific location in shortest time, there is no formal method to check the optimality of selected path, rather, he will decide by heuristics and based on his experience and knowledge about different paths and their traffic and waiting times in different periods of time. However, this type of selection can be improved by approximation using extended fuzzy logic and verified using the concept of f-validity presented by Zadeh in [35].

In real-world applications, there are context-dependent temporal and spatio-temporal uncertainties. Again, we use the map shown in Figure 3 to represent the complexity of temporal reasoning. We can observe that several ways exist for travelling from, say, Eiffel tower to Trocadero. It can be walked in 10-15 minutes, passing the bridge toward the destination. This will be selected only when group's preferred method of travelling inside a city is walking. Another way is using metro, which in this case may require a 10-15 minute walk from the tower to the nearest station, waiting for train to arrive, etc. This method takes more time in this special case, but will be the only selectable way if group's preference is strictly using metro. Therefore, the quality of mobile group decisions relies heavily on spatial and temporal contexts, which contain high-order uncertainty and require efficient context management to achieve reasonable results.

### 3.3.3. The human part of context

The human behaviour also plays an important role in context. Our needs and motivations are an significant source that set our goals, preferences and we select activities to achieve them. There are several psychological models for human needs. One of the well known models is Maslow's hierarchy of needs used in [36] for identifying user's requirements in service-oriented computing. The ERG theory [37] is another model which has extended and simplified Maslow's Hierarchy into a shorter set of three types of human needs: *Existence* (physical needs such as thirst and hunger), *Relatedness* (our social and friendship needs) and *Growth* (the need to grow, be creative for ourselves and for our environment and to achieve wholeness). Unlike Maslow's model, human needs in ERG theory do not form a hierarchy, but more of a continuum, in which several needs may exist simultaneously. In empirical studies, the ERG theory has achieved better results than Maslow's. Based on the ERG theory, if a higher-order need is frustrated, an individual may regress to increase the satisfaction of a lower-order need which appears easier to satisfy. This is known as the frustration-regression principle. We will use the ERG theory in several ways, combined with other aspects of context. For instance, monitoring the R-needs layer to improve our group-related actions including:

1. Filtering out the joining request from any user whose primary needs are of other types, and will degrade the group performance if joined. But we may have possible recommendation of other services to such users.

2. Monitoring each user's activity in terms of frequency of entering and leaving, say, R-needs layer, through a specific agent, making the service search space narrower down to those services which can fulfil this type of user needs.

In addition to the uncertainty that exists in definition of various types of needs, the degree of fulfilment for each type of needs and for each person, can be uncertain and requires uncertainty management methods.

# 4. A LAYERED DESIGN FOR f-CONTEXT

Based on the holistic LAP theory and our conceptual model of context, this section presents a layered architectural design for f-Context, using the CWW perceptual computer reasoning engine [38] which provides perceptual reasoning capability for type-2 fuzzy agents to perform complex qualitative reasoning tasks about context. The assumptions for our design and the scope of problem are as follows:

- A group or community of users with something in common, for example: a group of tourists, participants in a mobile auction, or members of a location-dependent social network [16]. They must use a common set of words, e.g. *near*, *far*, *very far* for contextual perceptions, with possible differences in defining the boundaries, i.e. fuzzy membership functions.

- A knowledgebase of personal and group rules based on type-2 fuzzy sets.

- Perceptions as words are input to our system. Inferring them and conversion from raw context is already studied by several researchers e.g. in [39, 40].

Our focus is on three complex parts of context as discussed in Sections 0 to 0. First, we give a set of design meta-rules based on LAP and ERG theories for managing uncertain context theories. The architecture will be elaborated in Section 0. The proposed architecture is platform dependent, but some guidelines for a sample specific implementation is given in Section o. We will present more detailed design description will be for the spatial part of context, which is extendable to other parts as well.

## 4.1. Meta-rules for our design

In [41] a high level foundation for perfect design of human-centric and communicative-based software is proposed. Their framework does not recommend any specific design, rather, having such a framework in mind, will motivate the designer to ask a suitable set of design questions which helps to address the "complex mixture of human needs, embodied in a weave of physical and social interactions". The idea is in "… designing spaces for human communication and interaction to expand those aspects of computing that are focused on people, rather than machinery" [41]. More specifically we should observe the following meta-rules:

- To apply the LAP approach to service-based context-aware access to information, we need to provide a mapping between human-initiated uncertain communication and existing crisp protocols for services [8].

- We need methods to handle an interactional form of context, rather than informational form [17], so our design should provide support for interaction between people through group communication, and perception of interactions as part of context-awareness.

Another set of design meta-rules comes from the EGR theory of human needs. In addition to using the ERG theory for modelling the individual human needs, it can be applied to group context. A well-managed group context should then have the following features:

- Provide a platform to fulfil the R-needs of its members by taking context into account.

- Manage long-term member satisfaction and provide relationship management with them, according to the higher levels of context in our LAP-based framework.

- Provide information services to as close to the needs of mobile people as possible, in a given geographic location, and with possible limitation in time since they are moving.

## 4.2. The f-Context architecture

To handle the context at a perceptual level, a candidate architecture is the perceptual computer [38] for implementing CWW based on type-2 fuzzy sets. This computer is capable of processing words. However, it does not process them directly and relies on some *mapping* from words to type-2 fuzzy sets (encoding) and vice-versa (decoding). The architecture of perceptual computer is described in detail by Mendel, et. al in references including [38, 42, 43]. The f-Context is designed as an service-based and agent-based architecture in which software agents use type-2 fuzzy reasoning capabilities (Figure 4). We describe each part of this architecture and its role briefly:

o Different types of context-related *perceptions* from various sources such as events and location sensing information are the inputs to f-Context. The only constraint on input words is that they must be already defined in vocabularies for which a *codebook* is created in f-Context. An interesting key process in any application of type-2 fuzzy logic like f-Context is the collaborative creation of these codebooks as suggested by Liu and Mendel in [44]. The process is called the Interval Approach (IA). In IA, every member of the group is asked to define an interval, usually between 0 and 10, for every linguistic variable such as *small*, *large*, *near* and *far*. Then all of these user-defined intervals are integrated into a single type-2 fuzzy set for each word. Thus, all group members participate in establishing the boundaries of each word for their set of common vocabulary. The resulting type-2 fuzzy set for each word is identified by its Footprint Of Uncertainty (FOU), which is a generalization of the type-1 fuzzy sets.

o *Persistence*, which comprises of the personal and group knowledgebase with different classes of rule-based group knowledge and context-specific codebooks.

o *Reasoning* is based on CWW using type-2 fuzzy sets, which in turn use the codebooks as the reference for reasoning tasks. The *decoder* converts the numeric results of reasoning from CWW engine to words as output, selected from an output vocabulary.

o *Autonomy*: Creating, maintaining and using the codebooks will require autonomic behaviours for managing them, to minimize the need for user intervention in internal f-Context processes. The required autonomous behaviour for extracting the contextual information from the communication between group members is provided through a set of agents, described in the following:

    o The *S-, T-, H-Agents* are responsible for keeping the corresponding codebook up-to-date and usable by CWW engine. For instance, the S-Agent keeps track of the S-Codebook containing FOUs for spatial context words which are created by our customized Interval Approach (IA) process as described later. Depending on the granularity of other context parts, location status of the group, new requirements emerged from member requests and other possible changes in the environment, the codebooks may require updating (re-creating) which is triggered by the *GroupManager* agent and followed by these agents.

    o The *GroupManager* is also responsible for some unique roles of a group, including the management of join and leave of group members, monitoring group context for changes, long-term relationship management and deployment-specific tasks. It can invoke collaborative codebook creation algorithm to build the membership functions for type-2 fuzzy sets by taking into account member's opinions, and uses them in operations like selecting people who can join the group. This agent also has the responsibility to monitor input queries, as a source of communicative act to

be used in a LAP-based way to detect further implicit actions. The usefulness of this approach will be shown in a sample scenario later.

o  The *PersonalAgents* exist for every member of the group, and are responsible for managing long- and short-term spatial, temporal and human needs context, perception of user's needs, mobility and possible crisp information. A personal knowledge base exists for every member of the group, and is updated by relevant events, including the fuzzy interval definition and participation of the user in building FOUs for handling inter-uncertainties.

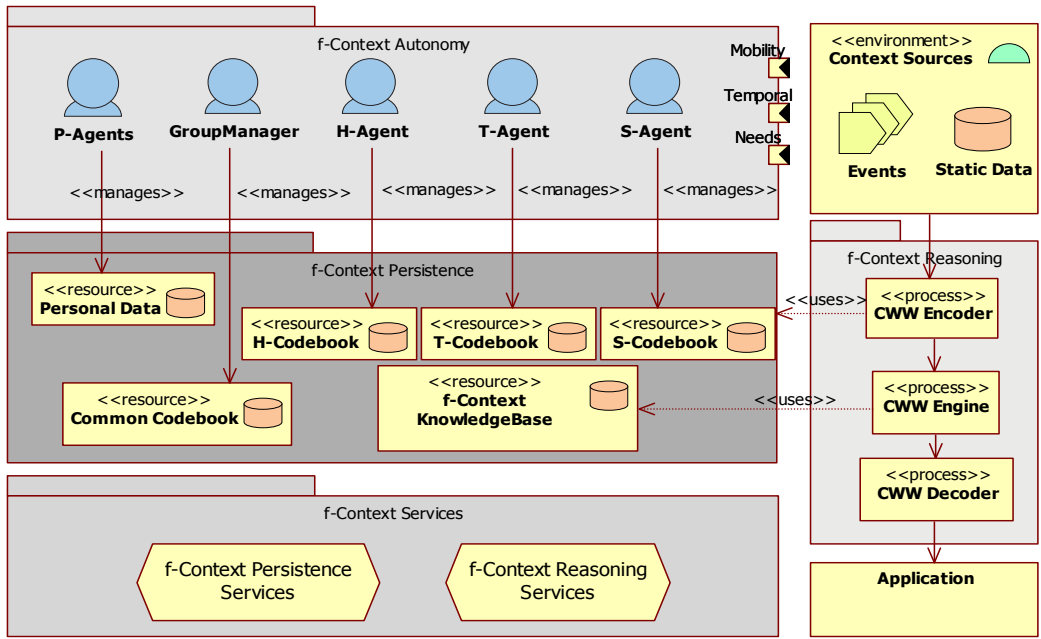

Figure 4. The high level architecture of f-Context.

o  *Context Services*: f-Context is designed to provide context-awareness functionality for ubiquitous information services with varying location in general, or for stationary services. The f-Context architecture is service-based itself, to provide platform independence and a discoverable interface.  Its functionality is exposed using a service-based architecture for extensibility and ease of use in various application scenarios. Examples of f-Context services are:

    o  Registering a rule or a set of rules in group knowledge base

    o  Selection of the appropriate type-2 fuzzy inference and decoding method for specific tasks

    o  Deployment-specific tasks for mobile or stationary services

    o  Settings and configuration tasks for f-Context

### 4.3. Using the f-Context architecture

Reasoning in f-Context is based on CWW. As mentioned in Section 0, each input or output of the reasoning process is a word, defined by its group-accepted type-2 fuzzy membership function, and stored in a codebook. In this section, we introduce some common sets for each of the three domains of our conceptual model of context in Section 3. Then we show that how a type-2 fuzzy knowledge base is built for managing the high-order uncertainty of such domains in a context-aware application.

### 4.3.1. Common vocabularies for contextual domains

For the spatial domain, a candidate vocabulary, which is the result of a survey on real-world web pages describing tourist Points Of Interest (POI), is adapted from [45]. They have collected a set of words for describing the position of a POI in relation to another POI, or in relation to a street and so on. This set is shown by $V_{S1}$ = {*Within walking distance, Across the street, Near, Close, Adjacent*}. For each of the words, i.e. the linguistic variables, one can easily observe that some inherent uncertainty exists with them, as discussed earlier in Section 3. Moreover, many of these spatial relationships mean different things to different people, and we are facing high-order uncertainty which is managed by collaborative creation and use of codebooks.

For the temporal domain, just like representing the space and spatial relationships, we also need a qualitative representation of time contextual reasoning with uncertainty. In fact, many real-world events have fuzzy beginning or ending. Handling temporal data and reasoning about time have been studied for decades. However, designing efficient methods of modelling temporal uncertainty still is an open research issue. In a recent work by [46], type-1 fuzzy logic is used for this purpose. They have used the well-known Allen's relationships and fuzzified them for reasoning. We can show the seven direct Allen's relationships by $V_{T1}$ set, which is a candidate vocabulary set for context perceptions given by $V_{T1}$ ={*Before, Overlaps, During, Meets, Starts, Finishes, Equals*}.

For the psychological human part of context, we also need a qualitative representation of human needs. An uncertain aspect of psychological human context is the trigger levels for transition from a specific type of need to another. The major inter-uncertainty exists here. For example one might assume a 60% fulfilment of needs in existence level is enough for the relationship needs to arise, while others may believe in 50% or 70%. A sample vocabulary for such transitional words can also be defined as, $V_{H1}$ = {*Existence_upward_fulfill, Relationship_upward_fulfill, Relationship_downward_fulfill, Growth_downward_fulfil*}.

Upward and downward words refer to the required fulfilment degree of current need type for the next type of need to arise. A larger set of transitions for Maslow's model can be defined as well. Our type-2 fuzzy rule-based model of context knowledgebase uses one or more of the above vocabulary sets to build antecedents for rules. The consequents must be also selected from proper vocabulary sets, depending on the application. For instance, if we are reasoning about the degree to which a location/service is important to visit/use, we can use a vocabulary set like

$V_{O1}$ = {*Unimportant, More or less unimportant, Moderately unimportant, More or less important, Moderately important, Very important*}.

This sample vocabulary was originally defined by [44] for investment decision-making domain. Another set can be defined for the degrees to which we rank a new candidate for joining a community. We show this candidate vocabulary set by $V_{O2}$ = {*Not recommended, More or less Recommended, Recommended, Highly recommended*}.

We emphasize that the output vocabularies for the consequents of fuzzy rules are highly application-dependent, but the same constraints for antecedent rules must be considered when selecting them.

### 4.3.2. The CWW knowledge-base in f-Context

The constructed codebooks are used by CWW engine for context-aware reasoning tasks. The internal mathematics of interval type-2 fuzzy reasoning is not discussed here and can be found in [42]. However, we will give an overview of it, to show the strength of type-2 fuzzy reasoning for solving some important context problems. The perceptual reasoning is performed in two steps. First, the firing interval of *each rule* is computed, so that one or more rules may fire to some degree depending on the input and FOUs in codebooks. Then the consequents of the *fired* rules will be combined using a Linguistic Weighted Average (LWA) method, so that each rule may affect the final result to some degree. To clarify the overall format of type-2 fuzzy rules which use the codebooks for reasoning, two sample applications for different mobile services are summarized in Table 1. Each service may require one or more parts of context, depending on which domains and which vocabulary sets affect the contextual group decision making and each part of context can affect the decision making process.

We observe that by having such powerful type-2 fuzzy reasoning capabilities, we can manage multi-criteria context reasoning, to take into account more than one part of uncertain context at a time. An illustrative example for usefulness of this method is described later in the verification section.

Table 1. Samples of using f-Context for mobile services.

| Service name | Ranking a mobile user to join the group | Nearby events check and rating |
|---|---|---|
| **Input context parts** | Spatial, Temporal, Needs | Spatial, Temporal |
| **Sample type-2 fuzzy rules for the member ranking service:**<br>  IF spatialProximity is *near* and needStatus is *Relationship_downward_fulfill*  THEN rank is *recommended*<br>  IF temporalStatus is *equals* and needStatus is *Relationship_downward_fulfill* THEN rank is *recommended*<br>  IF spatialProximity is *close*  and needStatus is *Relationship_upward_fulfill*  THEN rank is *highly recommended* | | |
| **Sample type-2 fuzzy rules for the event rating service:**<br>  IF eventLocation is *close* and eventTime is *before* THEN importance is *important*<br>  IF eventLocation is *near* and eventTime is *overlaps*  THEN importance is *more or less important*<br>  IF eventLocation is *far* and eventTime is *after*  THEN importance is *unimportant* | | |

It should be noted that combining fuzzy and non-fuzzy decision making may be required for optimum reasoning performance. As an example suppose using f-Context in a flight registration service when the desired flight must be before a specific event at a specific location. The reasoning for selection of the event can involve type-1 or type-2 fuzzy logic, but the interface with flight service calls for conventional exact reasoning. After all, we cannot reserve a flight on an uncertain time! Thus, a hybrid reasoning process will be required in many situations.

## 4.4. Verification of the design

In this section we verify our design from several aspects and then show how it can be applied in a real-world scenario. Since the design is based on perceptions and computing with words, it enables the context to be understandable and manageable at a higher and more expressive level. Different types of handling words as resulted from perceptions are possible. Examples of such perceptions are semantically enriched information about mobility from raw GPS data, perception of spatial and temporal information from news, events, user interactions, and so on. This can be regarded as a step toward true interactional context.

### 4.4.1. Relationship between context parts

Integration of CWW and agent-orientation into f-Context provides a sound basis for multi-criteria context reasoning capabilities. Examples of such uncertain context-aware perceptions, reasoning and actions in f-Context are:

1. Perception by *Mobility* perceptor that mobility_speed is slow, inferring that the vehicle_type of context is walk

2. Perception by *Mobility* perceptor that mobility_speed is medium, inferring that the vehicle_type of context is bicycle

3. Inferring from a long walk or long cycling (being percepted by *Mobility* preceptor) that some *E-Needs* may have risen in user, for example hunger

4. *Spatially* searching nearby providers (e.g. restaurants, news feeds, travel agents) who can fulfill these *needs*

5.  Inferring from a entering or approaching a new urban area that some *R-Needs* may have risen in user, for example finding new buddies

6.  *Spatially* searching nearby providers (people) who can fulfil this type of need, using a mobile matchmaker service, pre-filtering the people who have not similar *needs* part of context

7.  Communicating to establish a relationship

8.  Inferring from a medium or long relatedness activity that some *G-Needs* may have risen in both parties, offering some services that can fulfil this type of need, using a recommender service

### 4.4.2. Checking f-Context against the ERG theory

As can be seen from the architecture, personal agents continuously percept the users' contextual information, and monitor how services are used by them to fulfil the needs. Persistence of important interactional context operations will also help decision making in long-term patterns of context usage based on the history of contextual information.

### 4.4.3. Checking f-Context against LAP

Based on LAP, type-2 fuzzy sets defined as FOUs by group members can be considered a realization of *conversation for clarification* [7]. Using an example from patients and healthcare domain he has emphasized that "One can never guaranty that everything is totally precise. Precision is relative to each party's implicit anticipation that the other party will have a sufficiently *shared background* to carry out the action in a satisfactory way". Indeed, the proposed type-2 fuzzy managed context will provide powerful support for such a shared background between parties, by allowing them to share preferences in an uncertainty-aware, yet formally defined manner.

### 4.4.4. Security and privacy of contextual information in f-Context

The f-Context potentially provides several security and privacy preserving mechanisms, a vital need in today's mobile and service oriented systems [47, 48]. The agent-based architecture, also proposed in [49], means keeping information out of direct access by other users. Sensitive context information can be provided only after checking the closeness degree of the requester's context to current user, ensuring a reasonable trade-off between sharing context information and privacy.

## 4.5. Implementation of applications using f-Context

Our effort in this section was to present the f-Context architecture at conceptual level, without any dependency on any specific implementation platform. However, we suggest a sample configuration which can be used for implementation. A good candidate for implementation of the autonomy layer is JADE multi-agent platform. JADE uses FIPA-ACL compliant messaging, and is supported by several add-ons including JADE-LEAP for lightweight deployment of agents on Java-enabled cell phones. In JADE, *S-, T-, H-Agents* can be designed by assigning the spatial, temporal and other codebook monitoring tasks to *cyclic* agent behaviours. The coordination between agents in f-Context autonomy layer is initiated and led by the *GroupManager* agent, which has its own *one-shot* behaviour for creation of codebooks, and cyclic behaviours for monitoring and coordination tasks. One way to develop the personal *P-Agents* is using JADE-LEAP to deploy each personal agent on user device. Their main behaviours are managing contextual perceptions such as mobility and changes in user needs. They also communicate with *GroupManager* to perform the user interaction required for the first phase of IA process for defining type-2 fuzzy sets. The IA process is implemented by the authors of [44] in MATLAB, and can be invoked by the autonomy layer both for creation and usage of codebooks.

To implement the f-Context interface services, the Web Service Integration Gateway (WSIG) add-on for JADE can be used which allows automatically exposing agent services registered with the JADE's as web services. For the persistence layer, which is also responsible for providing an information sharing platform for agents in f-Context, we can use the LighTS tuple space tool, backed by any relational database for persistence of codebooks and other crisp f-Context information.

## 5. A REAL-WORLD SCENARIO: THE TOURIST GROUP GUIDE

In this section we present a tourist group scenario in detail, to show the feasibility of our proposed f-Context architecture, in addition to the sample service usage as already shown in Table 1. Getting information on POI and searching for location-based services (LBS) are two of the most popular mobile information services for tourists [50] and we will also show the usefulness of f-Context for exploiting such services by a group.

Consider a tourist group, with their group context hosted on certain server(s), travelling all around the world, keeping and enhancing their norms (actually stored in codebooks) as going on. New users may join at each location by an advertise/join approach, using the appropriate mechanisms presented in our design. For the specific scenario, we return to the location in Figure 3. Suppose that a member in of tourist group suggests "let's visit Eiffel tower today". The group mobile services platform, being supported by f-Context is triggered and engaged in the following order:

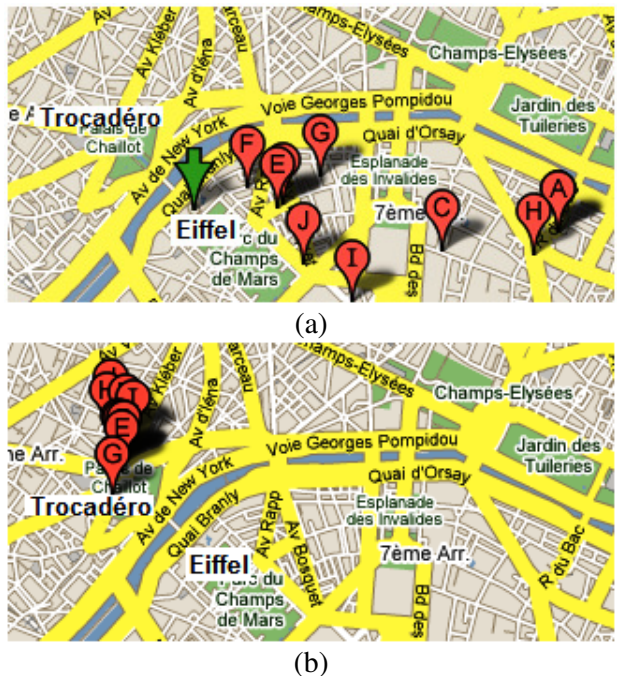

Figure 5. Google maps search for restaurants near (a) Eiffel tower; (b) Trocadero.

1) Looking for services to realize the main query, i.e. travelling to Eiffel tower, depending on current context of vehicle type.
2) Perception from the main query, by the group observer role and in a LAP-based way, that the group will need lunch after the visit. This inference can be based on current time (say, 10 am) and the group norm for the preferred duration of visiting a monument (say, 2 hours) predicting that the visit will finish about noon.

3) As a prerequisite for the next reasoning step, a request for setting norms can be sent to the members. The group context will determine by using type-2 fuzzy inference that whether, say, Trocadero should be considered close to Eiffel tower or not (see Figure 3).

4) Trying to look for a nearby restaurant, based on uncertain spatial reasoning and with regard to current context. If the group norms, defined by type-2 fuzzy rules, accept that Eiffel tower and Trocadero are counted as being *near* each other, then the results of both searches in Figure 5 can be combined. Although the current version of Google maps does not return any of the restaurants in Figure 5 –(b) for our query of (a), and vice versa, but we can use existing spatial uncertainty management techniques combined with our f-Context, to get a more acceptable result including nearest restaurants from the union of (a) and (b).

5) Predicting, based on ERG theory, and with some degree of uncertainty that after lunch and having this E-Need fulfilled, it may be a good choice to recommend services to group that will satisfy some R-Need (e.g. meeting with another tourist group) or G-Need.

## 6. RELATED WORK

In this section, we will review some important related works, categorized in several subsections, in addition to those referenced earlier in text. For the purpose of comparing with our work, we have selected a set of well-known architectures with different theoretical and architectural designs. This is not an exhaustive list of all context-aware frameworks, and the interested reader is referred to other references. A survey of different architectures for context and the role of communication in forming the context was initiated by Winograd in [6]. He classifies the architectures for context-aware systems to widget-based, service-based and blackboard systems. He argues that the general definition of context by Dey [51] needs to be more specific with more focus on human communication which contains uncertainty. A recent comprehensive survey and classification of context-aware architectures is performed by Hong, et al in [4]. In [5] the results of a another survey on several context-aware frameworks are presented. Most surveyed frameworks do not consider the dynamic and interactional view of context and do not deal with uncertainties, except the Gaia project [32] which uses probabilistic and type-1 fuzzy logic models. However, the group-context problems are not addressed by Gaia. The results of comparing the proposed f-Context architecture with the aforementioned approaches are summarized in Table 2.

Table 2. Comparing context-aware architectures.

| Framework | Theoretical Foundations | | | Architectural Aspects | | |
|---|---|---|---|---|---|---|
| | Holistic theory | Group-based | Uncertainty Management | Layered | Agent-based | Service-based |
| Gaia [32] | No | No | Yes, probabilistic and type-1 fuzzy sets | No | No | No |
| JCAF [52] | No | No | No | Yes | No | Yes |
| SAMOA [16] | No | Yes | No | Yes | No | Yes |
| Liao [53] | No | No | Yes, Dempster-Schafer | No | No | No |
| SCaLaDE [54] | No | No | No | Yes | Yes | Yes |
| SOCAM [55] | No | No | No | Yes | No | Yes |
| f-Context | Yes | Yes | Yes, CWW based on type-2 fuzzy sets | Yes | Yes | Yes |

The JCAF (Java Context-Aware Framework) [52] is a well-known Java-based framework implemented as a set of APIs for creating context-aware applications. It has a service-based architecture consisted of three layers for context sensors (acquiring raw context data), context service and context client. It does not follow any specific theory and is not designed to handle uncertainty. The role of context in social networks and location-dependent communities is considered for designing the Socially Aware and Mobile Architecture (SAMOA) architecture in [16]. They emphasize the group-based aspect of context, but their approach does not deal with individual or group uncertainty. A framework based for managing context information based on knowledge management is suggested by Liao, et al in [53] and illustrated using an m-commerce scenario. Their framework focuses on location data and uses ontology for knowledge representation. They use Dempster-Shafer evidence theory for reasoning and fusion of context sources. However, their framework is not designed to cope with high-order uncertainty and does not take other qualitative parts of context information such as the temporal and human need parts. They have also extended their context architecture with agents in [56]. The SCaLaDE (Services with Context awareness and Location awareness for Data Environments) architecture proposed by Bellavista in [54] is a service-based context-awareness middleware emphasizing the service management and the required policies. They also use mobile agents to support the required operations. However, their approach does not follow a specific theoretical foundation like most of the aforementioned approaches and their focus is on location context. SOCAM (Service-Oriented Context-Aware Middleware) is also based on an open infrastructure for context-awareness. It does not provide uncertainty management and autonomic, agent based context processing.

# 7. CONCLUSION

In this paper, we based our work on strong theories to design a context architecture for qualitative reasoning in context-aware services. To make this possible, we presented the idea in a step-by-step approach as summarized below:

1. We positioned context-awareness in a larger holistic view based on the LAP theory, to make it usable in as much other service-based works as possible.

2. Since LAP focuses on communication, and the communication contains uncertainty resulted from being performed by human, we proposed the agent-based f-Context model, based on CWW and type-2 fuzzy sets, to process contextual perceptions as words that are used in describing events, performing communication between members of a group, and in defining group knowledge base.

3. We defined a conceptual context model, which differentiates between crisp or simpler uncertain parts with more complex spatial, temporal and psychological parts, with a detailed formal view for building candidate vocabularies for spatial, temporal and human need domains. This model is of great importance for communities of mobile service providers and consumers. Mobility can be seen as changes in space (spatial context) and time (temporal context) initiated by the user needs.

4. We provided a layered, service-based architecture for implementing f-Context, verified through a step-by-step usage scenario for using f-Context in a tourist group application.

We believe that the concepts and architecture discussed in this paper can form a good foundation for thinking about context-awareness from a higher theoretical level and for future works on more realistic and human-centric group-based context-aware computing.

## Authors

**Nasser Ghadiri** is a PhD candidate of computer engineering at the Faculty of Engineering of the University of Isfahan (UI). He earned his M.Sc and B.Sc degrees from the University of Shiraz and Isfahan University of Technology, respectively. His research interests are spatial and mobile databases, computational intelligence and service-oriented architectures. He is a member of IEEE and ACM.

**Ahmad Baraani-Dastjerdi** is an assistant professor of computer engineering at the School of Engineering of the University of Isfahan (UI). He got his BS in Statistics and Computing in 1977. He got his MS & PhD degrees in Computer Science from George Washington University in 1979 & University of Wollongong in 1996, respectively. He is Head of the Research Department of the Communication systems and Information Security (CSIS) and Head of the ACM International Collegiate Programming Contest (ACM/ICPC) of University of Isfahan from 2000 until present. He co-authored three books in Persian and received an award of "the Best e-Commerce Iranian Journal Paper" (2005). Currently, he is teaching PhD and MS courses of Advance Topics in Database, Data Protection, Advance Databases, and Machining Learning. His research interests lie in Databases, Data security, Information Systems, e-Society, e-Learning, e-Commerce, Security in e-Commerce, and Security in e-Learning.

**Nasser Ghasem-Aghaee** is a professor of computer engineering at the Faculty of Engineering of the University of Isfahan (UI) and Sheikh-Bahaei Universty. He earned his PhD & MSc degrees from the University of Bradford and Georgia Tech, respectively. He spent two sabbatical leave (1993-94 & 2002-03) at the Ottawa Center of the McLeod Institute of Simulation Sciences, at Computer Science Department of the University of Ottawa, Ottawa, Ontario, Canada. He served as his Department Chair and Research and Graduate Studies Deputy Manager of Engineering College at the University of Isfahan between 1987 and 1993 and From 1994 until now, respectively. He authored three books in Persian and published more than 70 documents. He has been active in seminars and conferences held in different countries. His research interests have been in areas of Computer Simulation, Object-Oriented Analysis and Design, Artificial Intelligence (AI) and Expert Systems, AI in Software Engineering, AI in Simulation, OO in Simulation, AI in Object-Oriented Analysis, User Modelling, Advance Artificial Intelligence, and Software Agents and Applications.

**Mohammad A. Nematbakhsh** is an associate professor of computer engineering at the School of Engineering of the University of Isfahan (UI). He received his BSc in Electrical Engineering from Louisiana Tech University, USA, in 1981 and his MSc & PhD degrees in Electrical and Computer Engineering from University of Arizona, USA, in 1983 & 1987, respectively. He has published more than 70 papers and 3 US patents, and authored a book on database systems that is widely used in universities. He has received five awards and was the chair of the 6th CSI Computer Engineering Conference in 2001. He has been distinguished research fellow at the University of Isfahan and he was also awarded as the best national thesis advisor. He is the member of editorial board of several journals in Engineering Sciences. His main research interests include multi-agent systems applications in e-commerce and computer networks.